\documentclass{article}

\usepackage{arxiv}

\usepackage[utf8]{inputenc} 
\usepackage[T1]{fontenc}    
\usepackage{hyperref}       
\usepackage{url}            
\usepackage{booktabs}       
\usepackage{amsfonts}       
\usepackage{nicefrac}       
\usepackage{microtype}      
\usepackage{lipsum}
\usepackage{graphicx}
\graphicspath{ {./images/} }

\usepackage[utf8]{inputenc} 
\usepackage[T1]{fontenc}    
\usepackage{hyperref}       
\usepackage{url}            
\usepackage{booktabs}       
\usepackage{amsfonts}       
\usepackage{nicefrac}       
\usepackage{microtype}      
\usepackage{xcolor}         

\usepackage{graphicx}
\usepackage{algorithm}
\usepackage{amsmath}
\usepackage{amssymb}
\usepackage{float}
\usepackage{pifont}
\usepackage{algpseudocode}
\usepackage{natbib}
\usepackage{hyperref}
\usepackage{multirow}
\usepackage{amsthm}

\usepackage{textcomp}
\usepackage{caption}   
\usepackage[table]{xcolor}  

\title{\textbf{SO-Mamba}: \textbf{S}tate-\textbf{O}wnership Mamba for Unrolled MRI Reconstruction}

\author{
Pengcheng Fang\textsuperscript{1, *},
Hongli Chen\textsuperscript{2, *},
Fangfang Tang\textsuperscript{2},
Feng Liu\textsuperscript{2},
Xiaohao Cai\textsuperscript{1},
Shanshan Shan\textsuperscript{3, \textdagger}
\\[0.8em]
\textsuperscript{1}University of Southampton \quad
\textsuperscript{2}University of Queensland \quad 
\textsuperscript{3}Soochow University \quad
\\[0.4em]
\textsuperscript{\textdagger}Corresponding author.
}

\begin{document}

\maketitle

\begin{abstract}
Accelerated MRI reconstruction requires recovering missing details while preserving anatomically coherent structures across large spatial regions. State-space models such as Mamba provide efficient long-range modeling, making them attractive learned regularizers for unrolled reconstruction. However, in a data-consistency-coupled unrolled solver, different stages operate on different reconstruction iterates, where the resident carrier should preserve coherent reconstruction content across stages while stage-dependent non-resident evidence is tied to the current update. Treating these roles uniformly can place persistent resident-carrier evidence and update-dependent non-resident evidence into the same recurrent content route. We therefore propose SO-Mamba, a state-ownership Mamba regularizer that assigns reconstruction evidence within each Mamba stage to recurrent residency, state-interface access, and non-state output correction. SO-Mamba implements this ownership rule with a State-Ownership Router (SOR), which constructs a resident carrier for recurrent content and routes non-resident evidence to affine modulation of the B/C state interfaces and an output correction outlet. The resident carrier supplies the Mamba content route, while the non-resident evidence stream adapts the state interfaces and contributes through the output outlet without entering the recurrent content route. We further introduce a two-level outer-band leakage diagnostic that separates hidden-state storage from readout expression by measuring outer-band energy in
the selective-scan state trajectory and the post-scan Mamba readout. Experiments on five public MRI reconstruction benchmarks spanning diverse anatomies, sampling patterns, and coil configurations show that SO-Mamba consistently improves over CNN-, Transformer-, and Mamba-based baselines with competitive computational efficiency.
\end{abstract}

\section{Introduction}

Magnetic resonance imaging (MRI) is clinically indispensable because of its non-invasive acquisition
and superior soft-tissue contrast. However, long acquisition times increase patient discomfort
and limit clinical throughput~\citep{zbontar2018fastmri,shan2024b0inhomogeneity}. Accelerated MRI reduces
scan time by reconstructing images from undersampled k-space measurements, but undersampling
omits measurements and introduces aliasing artifacts~\citep{ye2019compressed}. Effective reconstruction
should therefore recover missing details while preserving anatomically coherent content under measurement
consistency.

Deep learning has advanced accelerated MRI reconstruction by learning
data-driven priors, from direct CNN mappings on zero-filled inputs~\citep{wang2016accelerating}
to learned modules in model-based and cascaded reconstruction frameworks~\citep{schlemper2017deep,qin2019crnn,hammernik2018variational}.
Such methods couple learned priors with the measurement model through
data-fidelity terms or explicit data-consistency updates. We focus on the
DC-coupled unrolled setting, where each learned regularizer is followed by a
data-consistency step, so each stage must refine the current iterate while
remaining compatible with acquired k-space measurements.

CNN regularizers are effective local operators, but their limited receptive fields hinder long-range
anatomical modeling under high acceleration and large fields of view. Vision Transformers improve
global context modeling through self-attention~\citep{dosovitskiy2020image}, but quadratic complexity limits
scalability to high-resolution clinical inputs~\citep{you2021unsupervised,wang2023dual}. Selective
state-space models, Mamba and its successors~\citep{gu2023mamba, lahoti2026mamba}, offer linear-complexity long-range modeling,
making them attractive regularizers for unrolled MRI reconstruction.

When Mamba is used as a regularizer in a data-consistency-coupled unrolled solver, efficiency alone does not determine how reconstruction evidence should be assigned inside the state-space computation. Each stage constructs a state from the current reconstruction iterate, while cross-stage information is updated through reconstruction iterates and DC steps rather than maintained in a shared Mamba hidden state. Assigning resident-carrier evidence and stage-dependent non-resident evidence to the same recurrent content route may therefore entangle roles that should be handled differently. The resulting question is which evidence should be allowed to become recurrent content at each stage.

We refer to this stage-local allocation rule as \emph{state ownership}. It separates three functional roles inside a Mamba regularizer: recurrent residency, state-interface access, and non-state output correction. The resident carrier receives recurrent residency by supplying the Mamba content route, whereas stage-dependent non-resident evidence adapts the state interfaces and corrects
the output without entering the recurrent content route. State ownership, therefore, specifies which evidence becomes recurrent content, which evidence controls how that content is written and read, and which evidence corrects the stage output through a non-state route.

We instantiate this principle as \textbf{SO-Mamba}, a \textbf{S}tate-\textbf{O}wnership Mamba regularizer for unrolled MRI
reconstruction. SO-Mamba uses a \textbf{State-Ownership Router} (\textbf{SOR}) to construct two streams: a resident carrier for recurrent content and a non-resident evidence stream for interface modulation and output correction. This implements the ownership rule within each stage-local regularizer while keeping non-resident evidence outside the recurrent content route. 

Our contributions are fourfold:





\begin{itemize}
    \item We identify stage-local state allocation as a distinct design problem
    in Mamba-based unrolled MRI reconstruction and formulate it as
    \emph{state ownership}.

    \item We propose SO-Mamba, a state-ownership regularizer built on Mamba-3 that assigns reconstruction evidence to recurrent residency, state-interface access, and
    non-state output correction within each unrolled reconstruction stage.

    \item We introduce SOR, a State-Ownership Router that constructs a resident carrier for recurrent content and routes non-resident evidence to
    interface modulation and output correction.
    
    \item We evaluate SO-Mamba on five MRI reconstruction benchmarks spanning
    diverse anatomies, sampling patterns, and coil configurations, and analyze
    its ownership behavior with a two-level diagnostic that separates hidden-state
    storage from readout-level expression.
\end{itemize}


\noindent\textbf{Unrolled MRI Reconstruction.}
Deep learning has been widely used for accelerated MRI reconstruction by learning data-driven priors. CNN-based methods directly map undersampled inputs to artifact-reduced images~\citep{jin2017deep,hyun2018deep}. Model-based deep
learning unrolls iterative optimization and combines learned regularization with data-fidelity terms or constraints derived from the measurement model, as in ADMM-Net~\citep{sun2016deep}, ISTA-Net~\citep{zhang2018ista}, and DuDoRNet~\citep{zhou2020dudornet}. In MRI reconstruction, data fidelity is often implemented through explicit k-space data-consistency updates that enforce agreement with acquired measurements. We focus on this DC-coupled setting and
study the learned regularizer from a state-allocation perspective, asking how reconstruction evidence should be assigned inside each reconstruction stage.

\noindent\textbf{Long-range Reconstruction Priors.}
CNN regularizers are effective local operators but limited in modeling
long-range anatomical dependencies. Transformer-based methods, including
SwinMR~\citep{yun2023swinmr}, ReconFormer~\citep{guo2023reconformer}, and
FpsFormer~\citep{meng2025boosting}, improve global context aggregation through
self-attention, but their quadratic complexity restricts scalability to
high-resolution reconstruction. Mamba~\citep{gu2023mamba} provides efficient
long-range modeling through selective state-space transitions, motivating recent
MRI reconstruction methods such as MambaMIR~\citep{zhao2024mambamir},
LMO~\citep{li2025lmo}, and HiFi-Mamba~\citep{chen2025hifi}. SO-Mamba builds on this line of Mamba-based reconstruction, but addresses a distinct stage-local state-assignment problem inside DC-coupled unrolled regularizers.

\section{Method}
\label{sec:method}

\subsection{Unrolled Reconstruction with Stage-Local Mamba States}
\label{sec:stage_local_allocation}

Accelerated MRI reconstruction estimates an image $x$ from undersampled
k-space measurements:
\begin{equation}
    y = A x + \epsilon ,
\end{equation}
where $A$ denotes the MRI forward operator, including Fourier encoding, the
k-space sampling mask, and coil sensitivity encoding when multi-coil data are
used. Let $x_k$ be the reconstruction estimate at the $k$-th unrolled stage. We
use an unrolled reconstruction formulation that alternates learned
regularization with data consistency:
\begin{equation}
    z_k = x_k + R_{\theta,k}(x_k),
    \qquad
    x_{k+1} = \mathrm{DC}(z_k, y, A),
\label{eq:unrolled_update}
\end{equation}
where $R_{\theta,k}$ is the learned regularizer at stage $k$, and
$\mathrm{DC}(\cdot)$ enforces agreement with the acquired measurements.

In a Mamba-based unrolled regularizer, each stage constructs a selective-scan
state from the current reconstruction iterate. This state is stage-local:
SO-Mamba does not propagate Mamba hidden states across unrolled stages. Across
stages, the reconstruction is updated through the sequence of regularization and
DC updates. The resulting allocation problem is which evidence should become
recurrent content within each stage-local regularizer, and which evidence should
instead control the state interfaces or correct the output.

\begin{figure*}[t]
    \centering
    \includegraphics[width=1.00\linewidth]{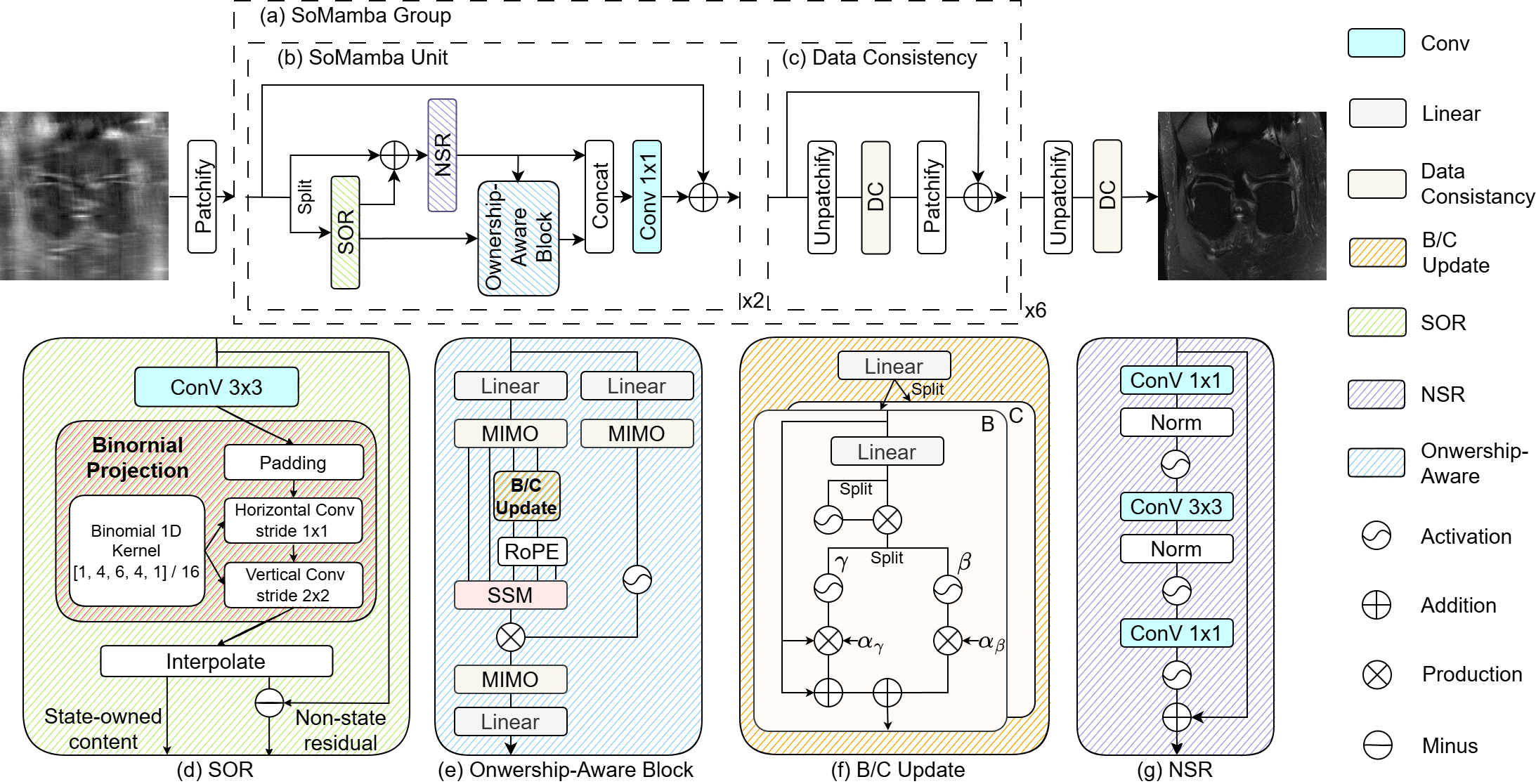}
    \caption{
    Overview of the SO-Mamba architecture.
    (a) The full network stacks six SO-Mamba groups; each group contains two
    SO-Mamba units followed by a data-consistency (DC) update.
    (b) Each SO-Mamba unit uses SOR to form a resident carrier and a non-resident evidence stream, processes them with Ownership-Aware blocks, and merges the output with a $1{\times}1$ convolution.
    (c) The DC block enforces measurement fidelity in $k$-space.
    (d) SOR forms $L_k$ and $G_k$ through fixed binomial carrier projection and off-carrier evidence routing.
    (e--g) The resident carrier supplies the Mamba content tokens, whereas the non-resident evidence stream modulates the $B/C$ state interfaces and contributes through the NSR output outlet.
    }
    \label{fig:arch}
\end{figure*}

\subsection{State Ownership as Path Assignment}
\label{sec:state_ownership}

We formulate the above allocation rule as \emph{state ownership}. A stage-local
Mamba regularizer exposes three routes through which evidence can affect the
reconstruction.

First, the content route produces the tokens scanned by Mamba. Evidence entering
this route receives recurrent residency, because it becomes the direct content
of the selective-scan state-space computation. Second, the interface route
controls how the state is written and read through the input-dependent $B$ and
$C$ components. Evidence entering this route receives state-interface access.
Third, the output route contributes to the reconstruction update outside the
recurrent content path. Evidence entering this route receives an output outlet.

For an evidence stream $E$, these three roles can be summarized as
\begin{equation}
\begin{gathered}
    \mathrm{residency:}\quad
    E \rightarrow u_k, \\
    \mathrm{access:}\quad
    E \rightarrow B'_k,\ C'_k, \\
    \mathrm{outlet:}\quad
    E \rightarrow Y_k
    \ \mathrm{through\ a\ non\text{-}state\ correction\ route},
\end{gathered}
\label{eq:state_roles}
\end{equation}
where $u_k$ denotes the Mamba content tokens, $B'_k$ and $C'_k$ denote the
conditioned state-space interfaces, and $Y_k$ denotes the output feature of the
regularization block.

SO-Mamba assigns these roles to two streams. Let $L_k$ denote the
resident carrier and $G_k$ denote the non-resident evidence stream at stage $k$.
The ownership contract is
\begin{equation}
\begin{gathered}
    L_k \rightarrow u_k,\qquad
    G_k \rightarrow B'_k,\ C'_k, \\
    G_k \rightarrow Y_k\ \mathrm{through\ an\ output\ outlet},\qquad
    G_k \nrightarrow u_k .
\end{gathered}
\label{eq:ownership_contract}
\end{equation}
Thus, $L_k$ receives recurrent residency, while $G_k$ receives state-interface
access and output correction. The non-resident evidence stream remains available for
reconstruction, but is excluded from the source of recurrent content tokens.

The exclusion $G_k \nrightarrow u_k$ restricts direct content residency rather
than all influence on the recurrent computation. Non-resident evidence still affects
the state-space path through the conditioned interfaces $B'_k$ and $C'_k$.
State ownership is therefore enforced as an architectural constraint: the
content, interface, and output routes are separated and assigned distinct
functional roles.

\subsection{SOR: State-Ownership Router}
\label{sec:sor}

The ownership contract in Eq.~\eqref{eq:ownership_contract} requires a concrete
routing mechanism that assigns reconstruction evidence to two different
privileges: recurrent residency and non-resident influence. SOR implements this
route assignment by producing a resident carrier $L_k$ and a non-resident
evidence stream $G_k$:
\begin{equation}
    (L_k, G_k) = \mathrm{SOR}(X_k),
    \label{eq:sor_overview}
\end{equation}
where $L_k$ is the only stream assigned to the Mamba content-token source, while
$G_k$ is routed to state-interface access and the non-state output outlet.

Given the current reconstruction estimate $x_k$, the learned regularizer first
extracts an image-domain feature tensor:
\begin{equation}
    X_k = \Phi_{\theta,k}(x_k),
    \label{eq:sor_feature}
\end{equation}
where $X_k \in \mathbb{R}^{C \times H \times W}$. SOR separates this
representation into two role pools:
\begin{equation}
    X_k =
    [X^{\mathrm{car}}_k, X^{\mathrm{nr}}_k],
    \label{eq:role_split}
\end{equation}
where $X^{\mathrm{car}}_k$ is used to define the resident content source and
$X^{\mathrm{nr}}_k$ provides non-resident evidence for state-interface access
and output correction. The two role pools are projected separately:
\begin{equation}
    U^{\mathrm{car}}_k =
    \phi_{\mathrm{car}}(X^{\mathrm{car}}_k),
    \qquad
    U^{\mathrm{nr}}_k =
    \phi_{\mathrm{nr}}(X^{\mathrm{nr}}_k).
    \label{eq:role_projection}
\end{equation}

\paragraph{Resident-carrier projection.}
SORP assigns recurrent residency through a fixed binomial carrier projector
$\mathcal{C}_{\mathbf{k}}$:
\begin{equation}
    L_k =
    \mathcal{C}_{\mathbf{k}}(U^{\mathrm{car}}_k),
    \label{eq:resident_carrier}
\end{equation}
where $L_k$ is the resident carrier returned at the original feature resolution.
The projector is channel-wise and non-adaptive, providing an architectural
eligibility rule for recurrent residency. Evidence that remains stable under
local neighborhood aggregation is used as the carrier, matching the role of
spatially coherent anatomical layout in the stage-local recurrent path.

The projection core uses a normalized 5-tap binomial kernel:
\begin{equation}
    \mathbf{k} =
    \frac{1}{16}[1,4,6,4,1],
    \label{eq:binomial_kernel}
\end{equation}
applied through separable horizontal and vertical depthwise passes with reflect
padding:
\begin{equation}
    \mathcal{P}_{\mathbf{k}}(U)
    =
    \mathrm{Conv}^{\mathrm{dw}}_{v}
    \Big(
        \mathcal{P}_{\mathrm{ref}}
        \big(
            \mathrm{Conv}^{\mathrm{dw}}_{h}
            (\mathcal{P}_{\mathrm{ref}}(U), \mathbf{k})
        \big),
        \mathbf{k}
    \Big),
    \label{eq:binomial_projection_core}
\end{equation}
where $\mathrm{Conv}^{\mathrm{dw}}_{h}$ and
$\mathrm{Conv}^{\mathrm{dw}}_{v}$ denote depthwise horizontal $1\times5$ and
vertical $5\times1$ convolutions, respectively, and
$\mathcal{P}_{\mathrm{ref}}$ denotes reflect padding. Because the projection is
depthwise, SORP does not mix channels before assigning resident and
non-resident roles. With kernel length $m=5$, the separable implementation has
cost $\mathcal{O}(2mHW)$ instead of $\mathcal{O}(m^2HW)$ for a full 2-D kernel.

In implementation, the carrier projector applies binomial projection, spatial
compaction, and resolution restoration:
\begin{equation}
    L_k =
    \mathcal{U}_{\times2}
    \left(
        \mathcal{D}_{\times2}
        \left(
            \mathcal{P}_{\mathbf{k}}(U^{\mathrm{car}}_k)
        \right)
    \right),
    \label{eq:carrier_projection_impl}
\end{equation}
where $\mathcal{D}_{\times2}$ denotes $2\times$ spatial compaction and
$\mathcal{U}_{\times2}$ denotes bilinear resolution restoration. This produces a
smooth resident carrier at the original feature resolution.

\paragraph{Non-resident evidence assignment.}
The carrier projector determines the part of the carrier-role pool assigned to
recurrent residency. The remaining carrier-rejected evidence is computed as
\begin{equation}
    E^{\mathrm{off}}_k =
    U^{\mathrm{car}}_k - L_k.
    \label{eq:carrier_rejected_evidence}
\end{equation}
This term captures local deviations from the resident carrier. SORP combines it
with the non-resident-role pool to form the residual evidence stream:
\begin{equation}
    G_k =
    U^{\mathrm{nr}}_k + E^{\mathrm{off}}_k .
    \label{eq:non_resident_stream}
\end{equation}
The stream $G_k$ remains available to the reconstruction, but it is not assigned
to the Mamba content-token source. Instead, it influences the recurrent path
through state-interface access and contributes through the non-state output
outlet. Thus, SORP instantiates Eq.~\eqref{eq:ownership_contract} by granting
recurrent residency to $L_k$ and non-resident influence to $G_k$.

\subsection{Ownership-Preserving Mamba-3 Update}
\label{sec:ownership_update}

Given the resident carrier $L_k$ and the non-resident evidence stream $G_k$, SO-Mamba
implements the ownership contract in the Ownership-Aware Mamba-3 block shown in
Fig.~\ref{fig:arch} (e). The resident carrier provides the recurrent content,
while the non-resident evidence stream adapts the $B/C$ state interfaces and contributes
through a non-state output outlet.

\textbf{Residency through the content route.}
Recurrent residency is enforced at the token source. The Mamba content tokens
are generated only from the resident carrier:
\begin{equation}
    u_k = \mathrm{Tokenize}(L_k).
\end{equation}
Thus, the resident carrier supplies the recurrent content route. The non-resident
evidence stream is excluded from the content-token source:
\begin{equation}
    G_k \nrightarrow u_k .
\end{equation}
This exclusion restricts only direct recurrent residency; $G_k$ can still affect
the state-space computation through the conditioned state interfaces.

\textbf{State-interface modulation.}
The non-resident evidence stream adapts the write and read interfaces of the state-space
computation. As shown in the $B/C$ update block in Fig.~\ref{fig:arch} (f),
SO-Mamba predicts affine modulation parameters from $G_k$:
\begin{equation}
    M_k =
    P(\mathrm{Norm}(G_k))
    =
    [\mu_B, \nu_B, \mu_C, \nu_C],
\end{equation}
where $P(\cdot)$ is a lightweight projection. The conditioned interfaces are
\begin{equation}
\begin{aligned}
    B'_k
    &=
    B_k \odot \big(1 + \alpha_{\mu}\tanh(\mu_B)\big)
    + \alpha_{\nu}\tanh(\nu_B), \\
    C'_k
    &=
    C_k \odot \big(1 + \alpha_{\mu}\tanh(\mu_C)\big)
    + \alpha_{\nu}\tanh(\nu_C),
\end{aligned}
\end{equation}
where $\odot$ denotes element-wise multiplication, and
$\alpha_{\mu}$ and $\alpha_{\nu}$ control the multiplicative and additive
modulation strengths. This gives $G_k$ state-interface access: it changes how
the resident content is written to and read from the state, without becoming a
source of recurrent content tokens.

\textbf{Mamba-3 state-space update.}
The conditioned interfaces are inserted into the Mamba-3 selective state-space
update. For notation, we first write a single-input single-output component.
Let $u_{k,n}$ denote the content token at position $n$, and define the
conditioned state input as
\begin{equation}
    v_{k,n} = B'_{k,n}u_{k,n}.
\end{equation}
Using the exponential-trapezoidal discretization of Mamba-3, the stage-local
recurrence is
\begin{equation}
    h_{k,n}
    =
    \alpha_{k,n}h_{k,n-1}
    +
    \beta_{k,n}v_{k,n-1}
    +
    \gamma_{k,n}v_{k,n},
\end{equation}
\begin{equation}
    s_{k,n}
    =
    {C'_{k,n}}^{\top}h_{k,n},
\end{equation}
with
\begin{equation}
    \alpha_{k,n}
    =
    \exp(\Delta_{k,n}A_{k,n}), \quad
    \beta_{k,n}
    =
    (1-\lambda_{k,n})\Delta_{k,n}\exp(\Delta_{k,n}A_{k,n}), \quad
    \gamma_{k,n}
    =
    \lambda_{k,n}\Delta_{k,n}.
\end{equation}
Here, $h_{k,n}$ is the stage-local recurrent state, $s_{k,n}$ is the state-space
readout, and $\lambda_{k,n}$ is the exponential-trapezoidal mixing coefficient.

In the implemented block, we use the Mamba-3 MIMO parameterization illustrated
in Fig.~\ref{fig:arch} (e). The ownership rule remains the same: the resident
carrier supplies the MIMO content inputs, while the non-resident evidence stream modulates
the corresponding $B/C$ interfaces. We denote the resulting post-scan readout as
\begin{equation}
    S_k =
    \mathrm{Mamba3}_{\mathrm{read}}
    (u_k; A_k, \Delta_k, \lambda_k, B'_k, C'_k).
\end{equation}

\textbf{Non-state output outlet.}
As shown in Fig.~\ref{fig:arch} (g), SO-Mamba implements the output outlet with
a Non-State Refinement (NSR) block applied to the non-resident evidence stream:
\begin{equation}
    Y_k =
    W_o\big([S_k, \mathrm{NSR}(G_k)]\big),
\end{equation}
where $[\cdot,\cdot]$ denotes channel-wise concatenation and $W_o$ is the output
projection. Thus, non-resident evidence contributes to the block output without
entering the recurrent content route.

The decoded reconstruction update is
\begin{equation}
    R_{\theta,k}(x_k) = \Delta x_k = \Psi_{\theta,k}(Y_k),
    \label{eq:decoded_update}
\end{equation}
which is inserted into the unrolled update in Eq.~\eqref{eq:unrolled_update}.

\subsection{Two-Level Outer-Band Leakage Diagnostic}
\label{sec:two_level_leakage_diagnostic}

We analyze the ownership behavior of SO-Mamba with a two-level outer-band
leakage diagnostic that separates recurrent-state trajectory content from
readout-level expression. The diagnostic measures outer-band energy on the
selective-scan hidden-state trajectory and the post-scan Mamba readout.

Let $h_{k,n}$ denote the stage-local recurrent state at token position $n$, and
let $S_k$ denote the post-scan, pre-output-projection Mamba readout. We reshape
the per-token hidden states back to the spatial token grid and flatten the
state, head, and rank dimensions as channels:
\begin{equation}
    H_k =
    \mathrm{Grid}\big(\{h_{k,n}\}_{n=1}^{N}\big)
    \in \mathbb{R}^{D_h \times H_t \times W_t},
    \label{eq:hidden_state_grid}
\end{equation}
where $N=H_tW_t$ and $D_h$ denotes the flattened hidden-state dimension. This
reshaped grid preserves the spatial correspondence of scan-indexed recurrent
states, but the states still reflect the dynamics of the selective-scan
trajectory. We therefore use hidden-state leakage as a trajectory-level
diagnostic rather than as a conventional spatial feature metric.

For a feature map $Z\in\mathbb{R}^{D\times H_t\times W_t}$ and a normalized
radial cutoff $r$, we define the outer-band leakage operator as
\begin{equation}
    \mathcal{L}_r(Z)
    =
    \frac{
        \sum_{d}
        \sum_{\rho(\xi) > r}
        \left|
            \mathcal{F}(Z_d)(\xi)
        \right|^2
    }{
        \sum_{d}
        \sum_{\xi}
        \left|
            \mathcal{F}(Z_d)(\xi)
        \right|^2
        +
        \varepsilon
    },
    \label{eq:leakage_operator}
\end{equation}
where $d$ indexes channels, $\mathcal{F}$ is the spatial Fourier transform,
$\xi$ indexes the 2-D transform grid, $\rho(\xi)$ is the normalized radial
distance from the grid center, and $\varepsilon$ is a small numerical constant.
We compute hidden-state leakage, readout leakage, and their expression ratio as
\begin{equation}
    \mathrm{HLeak}_r = \mathcal{L}_r(H_k), \qquad
    \mathrm{RLeak}_r = \mathcal{L}_r(S_k), \qquad
    \eta_r =
    \frac{\mathrm{RLeak}_r}{\mathrm{HLeak}_r+\varepsilon}.
    \label{eq:two_level_leakage}
\end{equation}

HLeak measures outer-band content present along the raw selective-scan state
trajectory, whereas RLeak measures the outer-band content expressed by the
post-scan readout after the state is read through the conditioned interface.
We use RLeak as the primary readout-level leakage metric, while HLeak indicates
whether outer-band components already emerge inside the recurrent state
trajectory. The ratio $\eta_r$ is reported as a readout-to-state expression
ratio, describing how much outer-band energy in the state trajectory is
expressed at the readout level. This diagnostic is not used for training; it
only analyzes how the recurrent path expresses outer-band content under
ownership ablations.

We use this diagnostic to analyze the three state roles in SO-Mamba. Removing
the non-state output outlet tests whether outer-band content is increasingly
expressed through the recurrent readout when the correction route is unavailable.
Removing state-interface access tests whether the recurrent path remains
carrier-dominant but loses adaptive read/write control, while parameter-tying
checks test whether leakage behavior is tied to the transition side
($A/\Delta$) or the read/write interfaces ($B/C$).

\section{Experiments}
\label{sec:experiments}

\subsection{Experimental Settings}
\label{sec:experimental_settings}

We compare SO-Mamba with UNet~\citep{ronneberger2015u}, ISTA-Net,
TransUNet~\citep{chen2021transunet}, ReconFormer,
FpsFormer, Mamba-UNet~\citep{wang2024mambaunet},
LMO, and HiFi-Mamba.
We evaluate SO-Mamba on five public MRI reconstruction benchmarks: fastMRI
knee~\citep{zbontar2018fastmri}, CC359 brain~\citep{warfield2004staple}, ACDC
cardiac~\citep{bernard2018deep}, M4Raw multi-coil brain~\citep{lyu2023m4raw}, and
Prostate158 prostate~\citep{adams2022prostate158}. These datasets cover diverse
anatomies, single- and multi-coil acquisitions, and different sampling settings.
More training, dataset, metrics, and preprocessing details are provided in \textbf{Appendix A}.




\subsection{Quantitative and Qualitative Results}
\label{sec:quantitative_results}

\begin{table*}[b]
\centering
\footnotesize
\setlength{\tabcolsep}{2.5pt}
\renewcommand{\arraystretch}{1.1}

\newcommand{\mstd}[2]{#1\raisebox{-0.35ex}{\tiny$\pm#2$}}
\newcommand{\bmstd}[2]{\textbf{#1}\raisebox{-0.35ex}{\tiny$\pm#2$}}

\caption{
Quantitative comparison on fastMRI and CC359 under equispaced Cartesian
sampling acceleration factors AF=4 and AF=8. Values are reported as
mean with standard deviation shown in smaller text.
}
\label{tab:mri_comparison}

\begin{tabular}{@{}l cccc cccc@{}}
\toprule
\multirow{3}{*}{\textbf{Method}}
& \multicolumn{4}{c}{\textbf{fastMRI}}
& \multicolumn{4}{c}{\textbf{CC359}} \\
\cmidrule(lr){2-5}
\cmidrule(lr){6-9}
& \multicolumn{2}{c}{\textbf{PSNR} $\uparrow$}
& \multicolumn{2}{c}{\textbf{SSIM} $\uparrow$}
& \multicolumn{2}{c}{\textbf{PSNR} $\uparrow$}
& \multicolumn{2}{c}{\textbf{SSIM} $\uparrow$} \\
\cmidrule(lr){2-3}
\cmidrule(lr){4-5}
\cmidrule(lr){6-7}
\cmidrule(lr){8-9}
& AF=4 & AF=8
& AF=4 & AF=8
& AF=4 & AF=8
& AF=4 & AF=8 \\
\midrule

Zero-Filling
& \mstd{29.25}{1.20} & \mstd{25.95}{1.18}
& \mstd{0.723}{0.049} & \mstd{0.620}{0.063}
& \mstd{24.79}{0.46} & \mstd{21.27}{0.38}
& \mstd{0.725}{0.013} & \mstd{0.576}{0.010} \\

\midrule

UNet
& \mstd{31.66}{1.60} & \mstd{28.60}{1.45}
& \mstd{0.798}{0.052} & \mstd{0.697}{0.070}
& \mstd{28.27}{0.60} & \mstd{24.28}{0.62}
& \mstd{0.847}{0.010} & \mstd{0.720}{0.014} \\

ISTA-Net
& \mstd{33.27}{1.85} & \mstd{29.44}{1.55}
& \mstd{0.832}{0.049} & \mstd{0.714}{0.071}
& \mstd{32.03}{0.70} & \mstd{25.44}{0.65}
& \mstd{0.902}{0.009} & \mstd{0.744}{0.014} \\

TransUNet
& \mstd{31.84}{1.65} & \mstd{29.01}{1.50}
& \mstd{0.810}{0.052} & \mstd{0.711}{0.076}
& \mstd{30.20}{0.70} & \mstd{25.69}{0.72}
& \mstd{0.890}{0.007} & \mstd{0.770}{0.015} \\

Mamba-UNet
& \mstd{33.68}{1.97} & \mstd{30.57}{1.62}
& \mstd{0.839}{0.047} & \mstd{0.742}{0.070}
& \mstd{31.08}{0.38} & \mstd{25.80}{0.57}
& \mstd{0.890}{0.006} & \mstd{0.761}{0.014} \\

ReconFormer
& \mstd{33.75}{1.95} & \mstd{30.42}{1.70}
& \mstd{0.837}{0.049} & \mstd{0.728}{0.071}
& \mstd{32.46}{0.70} & \mstd{26.47}{0.68}
& \mstd{0.906}{0.009} & \mstd{0.766}{0.015} \\

FpsFormer
& \mstd{33.74}{1.95} & \mstd{30.63}{1.70}
& \mstd{0.841}{0.049} & \mstd{0.732}{0.071}
& \mstd{32.35}{0.70} & \mstd{26.65}{0.68}
& \mstd{0.897}{0.009} & \mstd{0.768}{0.015} \\

LMO
& \mstd{34.49}{2.05} & \mstd{31.10}{1.75}
& \mstd{0.846}{0.048} & \mstd{0.744}{0.071}
& \mstd{35.35}{0.75} & \mstd{27.99}{0.72}
& \mstd{0.921}{0.008} & \mstd{0.787}{0.015} \\

HiFi-Mamba
& \mstd{34.47}{2.24} & \mstd{31.38}{1.85}
& \mstd{0.853}{0.048} & \mstd{0.758}{0.072}
& \mstd{35.74}{0.75} & \mstd{28.08}{0.81}
& \mstd{0.935}{0.009} & \mstd{0.802}{0.015} \\

\textbf{SO-Mamba}
& \bmstd{34.65}{1.91} & \bmstd{31.65}{1.43}
& \bmstd{0.855}{0.048} & \bmstd{0.762}{0.072}
& \bmstd{36.69}{0.76} & \bmstd{28.27}{0.74}
& \bmstd{0.954}{0.007} & \bmstd{0.830}{0.016} \\

\bottomrule
\end{tabular}
\end{table*}

\begin{table*}[h]
\centering
\footnotesize
\setlength{\tabcolsep}{3.5pt}
\renewcommand{\arraystretch}{1}

\providecommand{\mstd}[2]{#1\raisebox{-0.35ex}{\tiny$\pm#2$}}
\providecommand{\bmstd}[2]{\textbf{#1}\raisebox{-0.35ex}{\tiny$\pm#2$}}

\caption{
MRI reconstruction comparison across three datasets under different sampling settings.
Values are reported as mean with standard deviation shown in smaller text.
}
\label{tab:unified_mri_results}

\begin{tabular}{c l cc cc cc}
\toprule
\multirow{2}{*}{\textbf{AF}} 
& \multirow{2}{*}{\textbf{Method}}
& \multicolumn{2}{c}{\textbf{Prostate158}} 
& \multicolumn{2}{c}{\textbf{ACDC}} 
& \multicolumn{2}{c}{\textbf{M4Raw}} \\
\cmidrule(lr){3-4}
\cmidrule(lr){5-6}
\cmidrule(lr){7-8}
& 
& PSNR $\uparrow$ & SSIM $\uparrow$
& PSNR $\uparrow$ & SSIM $\uparrow$
& PSNR $\uparrow$ & SSIM $\uparrow$ \\
\midrule

\multirow{6}{*}{$\times 4$}
& UNet
& \mstd{25.96}{0.63} & \mstd{0.772}{0.020}
& \mstd{25.19}{1.20} & \mstd{0.796}{0.031}
& \mstd{30.02}{1.10} & \mstd{0.770}{0.030} \\

& TransUNet
& \mstd{25.45}{0.57} & \mstd{0.788}{0.027}
& \mstd{26.18}{1.25} & \mstd{0.813}{0.031}
& \mstd{30.59}{1.08} & \mstd{0.762}{0.031} \\

& Mamba-UNet
& \mstd{27.05}{0.67} & \mstd{0.776}{0.025}
& \mstd{27.58}{1.15} & \mstd{0.812}{0.029}
& \mstd{31.07}{0.93} & \mstd{0.777}{0.029} \\

& LMO
& \mstd{28.15}{0.80} & \mstd{0.808}{0.025}
& \mstd{28.74}{1.35} & \mstd{0.833}{0.030}
& \mstd{29.86}{1.57} & \mstd{0.769}{0.022} \\

& HiFi-Mamba
& \mstd{28.58}{0.86} & \mstd{0.827}{0.023}
& \mstd{32.41}{1.91} & \mstd{0.916}{0.027}
& \mstd{31.13}{0.82} & \mstd{0.778}{0.028} \\

& \textbf{SO-Mamba}
& \bmstd{29.07}{0.89} & \bmstd{0.840}{0.023}
& \bmstd{33.18}{1.96} & \bmstd{0.927}{0.026}
& \bmstd{31.62}{1.08} & \bmstd{0.791}{0.028} \\

\midrule

\multirow{6}{*}{$\times 8$}
& UNet
& \mstd{21.70}{0.48} & \mstd{0.584}{0.028}
& \mstd{22.21}{1.10} & \mstd{0.640}{0.041}
& \mstd{28.14}{0.90} & \mstd{0.724}{0.034} \\

& TransUNet
& \mstd{21.47}{0.44} & \mstd{0.565}{0.032}
& \mstd{22.69}{1.11} & \mstd{0.663}{0.041}
& \mstd{28.95}{0.90} & \mstd{0.728}{0.034} \\

& Mamba-UNet
& \mstd{22.35}{0.46} & \mstd{0.596}{0.028}
& \mstd{23.78}{1.09} & \mstd{0.660}{0.037}
& \mstd{29.43}{0.84} & \mstd{0.738}{0.033} \\

& LMO
& \mstd{22.68}{0.47} & \mstd{0.600}{0.030}
& \mstd{24.18}{1.18} & \mstd{0.677}{0.039}
& \mstd{28.35}{1.19} & \mstd{0.728}{0.026} \\

& HiFi-Mamba
& \mstd{23.49}{0.60} & \mstd{0.647}{0.036}
& \mstd{27.06}{1.49} & \mstd{0.792}{0.042}
& \mstd{29.65}{0.82} & \mstd{0.741}{0.032} \\

& \textbf{SO-Mamba}
& \bmstd{23.76}{0.65} & \bmstd{0.657}{0.035}
& \bmstd{27.45}{1.52} & \bmstd{0.804}{0.041}
& \bmstd{29.81}{0.81} & \bmstd{0.745}{0.032} \\

\bottomrule
\end{tabular}
\end{table*}

\vspace{3pt}\noindent\textbf{Quantitative Results.} Tables~\ref{tab:mri_comparison} and~\ref{tab:unified_mri_results} compare
SO-Mamba with CNN-, Transformer-, and Mamba-based baselines across five
datasets, covering equispaced Cartesian, random, radial, and multi-coil
settings. SO-Mamba achieves the best results on fastMRI and CC359 under both
AF=4 and AF=8, with representative AF=4 scores of 34.65\,dB/0.855 SSIM on
fastMRI and 36.69\,dB/0.954 SSIM on CC359. It also leads on Prostate158 random
sampling, ACDC radial sampling, and M4Raw multi-coil reconstruction, reaching
29.07\,dB/0.840, 33.18\,dB/0.927, and 31.62\,dB/0.791 at AF=4, respectively.
The AF=8 results follow the same trend, showing consistent generalization across
anatomies and acquisition settings.

\begin{figure*}[t]
    \centering
    \includegraphics[width=1.005\linewidth]{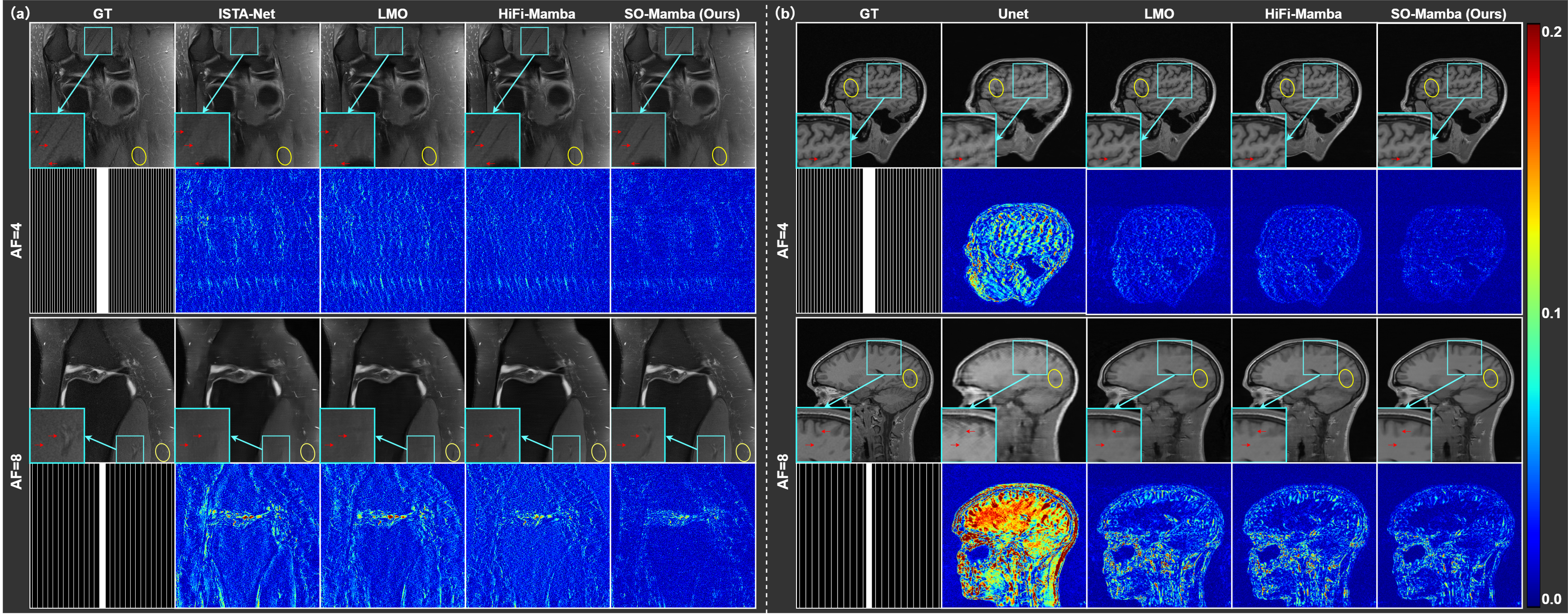}
    \caption{
    Qualitative comparison on fastMRI and CC359 under single-coil settings.
    (a) Reconstruction results on fastMRI with acceleration factors
    AF=4 and AF=8. (b) Reconstruction results on CC359 under the same
    acceleration factors. The second row of each subplot shows the corresponding error maps. Blue boxes, yellow ellipses, and red arrows highlight local reconstruction details.
    }
    \label{fig:result}
\end{figure*}

\vspace{3pt}\noindent\textbf{Qualitative Results.} Figure~\ref{fig:result} compares reconstructions on fastMRI and CC359 at acceleration factors AF=4 and AF=8. CNN-based baselines show blurred edges,
while Transformer- and Mamba-based baselines still exhibit boundary degradation
and local artifacts. SO-Mamba produces sharper tissue boundaries,
smoother intensity transitions, and more uniform error maps across both
anatomies.

\subsection{Efficiency Analysis}
\label{sec:efficiency}
We evaluate efficiency on fastMRI at $8\times$ acceleration with $320\times320$
inputs. As shown in Table~\ref{tab:efficiency_fastmri}, SO-Mamba achieves the
best PSNR/SSIM among compared efficient reconstruction models while using the
lowest FLOPs, improving over LMO and HiFi-Mamba in both accuracy and efficiency.

\begin{table*}[t]

\begin{minipage}[t]{0.35\textwidth}
\centering
\footnotesize
\setlength{\tabcolsep}{4pt}
\renewcommand{\arraystretch}{1.1}
\captionof{table}{
Efficiency comparison on fastMRI at $8\times$ acceleration.
}
\label{tab:efficiency_fastmri}
\begin{tabular}{l c c c}
\toprule
\textbf{Method}
& \textbf{PSNR} $\uparrow$
& \textbf{SSIM} $\uparrow$
& \textbf{FLOPs} $\downarrow$ \\
\midrule
ReconFormer
& 30.42 & 0.728 & 270.60G \\

FpsFormer
& 30.63 & 0.732 & 200.45G  \\

LMO
& 31.10 & 0.744 & 484.98G  \\

HiFi-Mamba
& 31.38 & 0.758 & 67.87G \\

\textbf{SO-Mamba}
& \textbf{31.65}
& \textbf{0.762}
& \textbf{45.59G}\\
\bottomrule
\end{tabular}
\end{minipage}
\hspace{0.33in} 
\begin{minipage}[t]{0.52\textwidth}
\centering
\footnotesize
\setlength{\tabcolsep}{1pt}
\renewcommand{\arraystretch}{0.95}
\captionof{table}{
State-ownership path ablation on fastMri at $4\times$ acceleration. 
}
\label{tab:ownership_ablation}

\begin{tabular}{l c c c c c c}
\toprule
\textbf{Variant}
& \textbf{SOR}
& $\boldsymbol{G_k\!\to\!u_k}$
& \textbf{Access}
& \textbf{Outlet}
& \textbf{PSNR} $\uparrow$
& \textbf{SSIM} $\uparrow$ \\
\midrule

Mamba regularizer
& -- & -- & -- & -- & 33.92 & 0.802 \\

+ SOR router
& \checkmark & -- & -- & -- & 34.38 & 0.852 \\

w/o state access
& \checkmark & -- & -- & \checkmark & 34.47 & 0.849 \\

w/o output outlet
& \checkmark & -- & \checkmark & -- & 34.34 & 0.854 \\

\rowcolor{gray!13}
w/ content residency
& \checkmark & \checkmark & \checkmark & \checkmark & \emph{31.71} & \emph{0.762} \\

\textbf{SO-Mamba}
& \checkmark & -- & \checkmark & \checkmark
& \textbf{34.65} & \textbf{0.855} \\

\bottomrule
\end{tabular}

\end{minipage}
\end{table*}

\subsection{Mechanism Analysis and Ablation}
\label{sec:mechanism_ablation}

The ablation study examines whether the proposed routes implement the
state-ownership contract. We analyze two functional routes for non-resident
evidence: state-interface access through the $B/C$ interfaces, and the
non-state output outlet outside the recurrent content route. We also test an
ownership-violating variant that allows $G_k$ to enter the Mamba content-token
route.

\vspace{3pt}\noindent\textbf{State-Ownership Ablation.}
Table~\ref{tab:ownership_ablation} shows that the ownership routes are
complementary. Removing state-interface access or the output outlet reduces
PSNR from 34.65\,dB to 34.47\,dB and 34.34\,dB, respectively. More importantly,
allowing non-resident evidence to obtain content residency sharply degrades
performance to 31.71\,dB, showing that feeding $G_k$ into the recurrent content
route violates the ownership contract. These results support our design:
$G_k$ should influence reconstruction through state-interface access and the
output outlet, rather than becoming a source of recurrent content tokens.

\vspace{3pt}\noindent\textbf{Two-Level Outer-Band Leakage.}
Table~\ref{tab:outer_band_leakage} separates raw hidden-state storage from
readout-level expression. HLeak measures outer-band energy in the selective-scan
hidden-state trajectory, RLeak measures the post-scan readout, and
$\eta=\mathrm{RLeak}/\mathrm{HLeak}$ measures how strongly outer-band content is
expressed by the recurrent readout.

Removing the output outlet sharply increases RLeak by 4.9$\times$/5.8$\times$,
with $\eta$ rising close to 1. This indicates that, when the non-state correction
route is unavailable, outer-band content is expressed through the recurrent
readout. In contrast, removing state-interface access lowers RLeak but also
reduces reconstruction quality, showing that low readout leakage alone is not
sufficient; non-resident evidence still needs controlled access to the state
interfaces. The tied $B/C$ variant remains close to SO-Mamba, while tied
$A/\Delta$ increases readout leakage, further supporting the separation between
recurrent residency, interface access, and output correction.

\begin{table*}[t]
\centering
\footnotesize
\caption{
Two-level outer-band leakage under ownership ablations and parameter-tying
sanity checks. HLeak is measured on the selective-scan hidden-state trajectory,
RLeak on the post-scan, pre-output-projection Mamba readout, and
$\eta=\mathrm{RLeak}/\mathrm{HLeak}$ measures readout-to-state outer-band
expression. Rel. reports the relative RLeak compared with SO-Mamba.
}
\label{tab:outer_band_leakage}
\setlength{\tabcolsep}{4.5pt}
\renewcommand{\arraystretch}{1.2}
\begin{tabular}{c c c c c c c c c c c}
\hline
\multirow{2}{*}{\textbf{Variant}}
& \multirow{2}{*}{\textbf{Access}}
& \multirow{2}{*}{\textbf{Outlet}}
& \multicolumn{4}{c}{$\boldsymbol{r=0.25}$}
& \multicolumn{4}{c}{$\boldsymbol{r=0.35}$} \\
\cline{4-7}
\cline{8-11}
&
&
& \textbf{HLeak}
& \textbf{RLeak}
& $\boldsymbol{\eta}$
& \textbf{Rel.}
& \textbf{HLeak}
& \textbf{RLeak}
& $\boldsymbol{\eta}$
& \textbf{Rel.} \\
\hline
\multicolumn{11}{c}{\textit{Ownership-route ablations}} \\
\hline
SO-Mamba
& \checkmark & \checkmark
& 0.2323 & 0.0229 & 0.098 & 1.0$\times$
& 0.1750 & 0.0143 & 0.082 & 1.0$\times$ \\

w/o output outlet
& \checkmark & --
& 0.1130 & 0.1122 & 0.993 & 4.9$\times$
& 0.0900 & 0.0824 & 0.916 & 5.8$\times$ \\

w/o state access
& -- & \checkmark
& 0.2137 & 0.0104 & 0.049 & 0.45$\times$
& 0.1618 & 0.0045 & 0.028 & 0.31$\times$ \\
\hline
\multicolumn{11}{c}{\textit{Parameter-tying sanity checks}} \\
\hline
tied $A/\Delta$
& \checkmark & \checkmark
& 0.2281 & 0.0399 & 0.175 & 1.7$\times$
& 0.1849 & 0.0273 & 0.148 & 1.9$\times$ \\

tied $B/C$
& \checkmark & \checkmark
& 0.2582 & 0.0227 & 0.088 & 0.99$\times$
& 0.2017 & 0.0140 & 0.069 & 0.98$\times$ \\
\hline
\end{tabular}
\end{table*}

\section{Conclusion and Limitations}
\label{sec:conclusion}

We presented SO-Mamba, a state-ownership Mamba regularizer for unrolled MRI
reconstruction. SO-Mamba assigns reconstruction evidence to recurrent residency,
state-interface access, and non-state output correction through SOR, which
constructs a resident carrier and routes non-resident evidence to access and
correction paths. Experiments on five MRI benchmarks show strong reconstruction
performance, and the two-level leakage diagnostic indicates that SO-Mamba
separates hidden-state storage from readout-level outer-band expression. SOR is
an architectural assignment rule rather than a semantic decomposition of anatomy
and artifact. Future work may explore adaptive ownership routing and broader
clinical acquisition settings.



\bibliographystyle{plainnat}
\bibliography{reference}
\appendix

\section{Detailed Experimental Settings}
\label{app:experimental_details}

\subsection{Data Preprocessing}
\label{app:processing}

The preprocessing for fastMRI and CC359 follows the data preprocessing pipeline
of HiFi-Mamba. For Prostate158, we adopt the same procedure, except that the
sampling mask is replaced with a random mask. For ACDC, we employ a golden-angle
radial sampling mask and first perform a center crop to an image size of
$128\times128$. Four-fold undersampling corresponds to using 32 radial spokes,
while eight-fold undersampling uses 16 spokes. For M4Raw, we use the equispaced
Cartesian undersampling mask and split the real and imaginary parts of the
four-channel $k$-space input to form an eight-channel input.

\subsection{Dataset Introduction}
\label{app:datasets}

The fastMRI dataset used in our study consists of 1,172 complex-valued
single-coil coronal knee examinations. Each volume contains approximately 35
coronal slices with an in-plane matrix size of $320\times320$. We train and
evaluate all models on the proton-density fat-suppressed (PDFS) subset,
strictly following the official train/validation/test split.

The CC359 dataset contains raw brain MR scans acquired on clinical MR scanners.
Following the official split, we use 25 subjects for training and 10 subjects
for testing. Each 2D slice has an in-plane matrix size of $256\times256$.

The ACDC dataset consists of cardiac MR images from 150 subjects. Following the
official split, we use 100 cases for training and 50 cases for testing, each
containing short-axis slices with an in-plane resolution of $256\times256$.

M4Raw contains multi-coil brain MRI $k$-space data acquired at 0.3T with a
four-channel coil. Following the official split, we use 384 T1-weighted volumes
for training and 150 T1-weighted volumes for testing.

Prostate158 provides prostate MR images from 158 volunteers, split into 139
training and 19 testing cases.

\subsection{Evaluation Protocol}
\label{app:evaluation}

We report Peak Signal-to-Noise Ratio (PSNR) and Structural Similarity Index
(SSIM) as the main reconstruction metrics. Both metrics are higher-is-better.
For each test case, we first compute the metric over valid slices and average
the slice-level scores within the case. We then report the mean and standard
deviation across test cases. Therefore, the standard deviations in the main
tables reflect case-level variability rather than random-seed variability.

For qualitative comparison, we show reconstructed images and corresponding
error maps under $4\times$ and $8\times$ acceleration. Error maps are visualized
with a fixed color range to make local reconstruction differences comparable
across methods.

\subsection{Training Configuration}
\label{app:training}

All reproduced models are trained with the AdamW optimizer. The initial learning
rate is set to $8\times10^{-4}$, followed by a cosine annealing schedule. A
five-epoch warm-up is used at the beginning of training. All models are trained
for 100 epochs.

Unless otherwise specified, the same training protocol is used for SO-Mamba and
the reproduced baselines. Experiments are conducted on two NVIDIA H100 GPUs.
FLOPs are measured on an NVIDIA A100 using inputs of size $320\times320$.

\begin{table}[h]
\centering
\footnotesize
\begin{tabular}{l|l}
\toprule
\textbf{Item} & \textbf{Setting} \\
\midrule
Optimizer & AdamW \\
Loss & L1 loss \\
Initial learning rate & $8\times10^{-4}$ \\
Learning-rate schedule & Cosine annealing \\
Warm-up & 5 epochs \\
Training epochs & 100 \\
Training GPUs & 2 NVIDIA H100 GPUs \\
FLOPs measurement & NVIDIA A100 \\
Input size for FLOPs & $320\times320$ \\
Main metrics & PSNR, SSIM \\
Reported deviation & Case-level standard deviation \\

\bottomrule
\end{tabular}
\caption{Training and evaluation configuration.}
\label{tab:training_config}
\end{table}

\subsection{Mamba-3 Configuration}
\label{app:mamba3_config}

We use Mamba-3 as the state-space backbone inside the learned regularizer. The
model dimension is set to $d_{\mathrm{model}}=96$. The selective state size is
set to $d_{\mathrm{state}}=16$, and the head dimension is set to
$d_{\mathrm{head}}=64$. We use the MIMO formulation with rank $4$ and chunk size
$16$. Output projection normalization is disabled. Settings are shown in
Table~\ref{tab:mamba3_config}.

\begin{table}[t]
\centering
\footnotesize
\begin{tabular}{l|c}
\toprule
\textbf{Hyperparameter} & \textbf{Setting} \\
\midrule
Model dimension $d_{\mathrm{model}}$ & 96 \\
State size $d_{\mathrm{state}}$ & 16 \\
Head dimension $d_{\mathrm{head}}$ & 64 \\
MIMO mode & Enabled \\
MIMO rank & 4 \\
Chunk size & 16 \\
Output projection norm & Disabled \\
\bottomrule
\end{tabular}
\caption{
Mamba-3 configuration used in the learned regularizer.
}
\label{tab:mamba3_config}
\end{table}

\subsection{Baselines}
\label{app:baselines}

We compare SO-Mamba with CNN-, Transformer-, and Mamba-based reconstruction
baselines. The CNN-based baselines include UNet and ISTA-Net. Transformer-based
baselines include TransUNet, ReconFormer, and FpsFormer. Mamba-based baselines
include Mamba-UNet, LMO, and HiFi-Mamba.

For reproduced baselines, we use the same data preprocessing, sampling masks,
training schedule, and evaluation protocol whenever possible. For methods with
official implementations or published configurations, we follow the reported
settings and adapt only the input/output interface required by the corresponding
dataset and sampling setting.

\subsection{Two-Level Outer-Band Leakage Diagnostic}
\label{app:leakage_details}

We compute outer-band leakage at two levels: the selective-scan hidden-state
trajectory and the post-scan, pre-output-projection Mamba readout. For
hidden-state leakage, the per-token recurrent states are reshaped back to the
spatial token grid, with state, head, and rank dimensions flattened as channels:
\begin{equation}
    H_k =
    \mathrm{Grid}\big(\{h_{k,n}\}_{n=1}^{N}\big)
    \in \mathbb{R}^{D_h \times H_t \times W_t},
\end{equation}
where $N=H_tW_t$ and $D_h$ denotes the flattened hidden-state dimension.

For a feature map $Z\in\mathbb{R}^{D\times H_t\times W_t}$ and a normalized
radial cutoff $r$, the leakage operator is
\begin{equation}
    \mathcal{L}_r(Z)
    =
    \frac{
        \sum_d
        \sum_{\rho(\xi)>r}
        |\mathcal{F}(Z_d)(\xi)|^2
    }{
        \sum_d
        \sum_{\xi}
        |\mathcal{F}(Z_d)(\xi)|^2
        +
        \varepsilon
    },
\end{equation}
where $d$ indexes channels, $\mathcal{F}$ is the spatial Fourier transform,
$\xi$ indexes the 2-D transform grid, and $\rho(\xi)$ denotes the normalized
radial distance from the grid center. We compute hidden-state leakage, readout
leakage, and their expression ratio as
\begin{equation}
    \mathrm{HLeak}_r =
    \mathcal{L}_r(H_k),
    \qquad
    \mathrm{RLeak}_r =
    \mathcal{L}_r(S_k),
    \qquad
    \eta_r =
    \frac{\mathrm{RLeak}_r}{\mathrm{HLeak}_r+\varepsilon}.
\end{equation}

In the main analysis, leakage is computed on the fastMRI $8\times$ validation
setting. We aggregate leakage over SO-Mamba units and slices within each case,
and report case-level averages. HLeak measures outer-band storage in the raw
selective-scan state trajectory, while RLeak measures outer-band expression in
the recurrent readout.

\subsection{Implementation Notes}
\label{app:implementation_notes}

All reported models are trained and evaluated with the same reconstruction
protocol unless otherwise specified. Data-consistency layers are implemented
according to the corresponding forward operator: single-coil Cartesian settings
use the single-coil Fourier sampling operator, while multi-coil settings include
coil sensitivity encoding. FLOPs are measured using the forward pass of the
reconstruction model at the specified input resolution and do not include data
loading or visualization overhead.

\end{document}